\documentclass[letterpaper]{article} % DO NOT CHANGE THIS
\usepackage{aaai2026}  % DO NOT CHANGE THIS
\usepackage{times}  % DO NOT CHANGE THIS
\usepackage{helvet}  % DO NOT CHANGE THIS
\usepackage{courier}  % DO NOT CHANGE THIS
\usepackage[hyphens]{url}  % DO NOT CHANGE THIS
\usepackage{graphicx} % DO NOT CHANGE THIS
\usepackage{natbib}  % DO NOT CHANGE THIS AND DO NOT ADD ANY OPTIONS TO IT
\usepackage{caption} % DO NOT CHANGE THIS AND DO NOT ADD ANY OPTIONS TO IT
\usepackage{algorithm}
\usepackage{algorithmic}

\usepackage{amsmath}
\usepackage{amssymb}
\usepackage{booktabs}
\usepackage{multirow} 
\usepackage{newfloat}
\usepackage{listings}
\DeclareCaptionStyle{ruled}{labelfont=normalfont,labelsep=colon,strut=off} % DO NOT CHANGE THIS
\floatstyle{ruled}
\newfloat{listing}{tb}{lst}{}
\floatname{listing}{Listing}
\title{REAP: Enhancing RAG with Recursive Evaluation and Adaptive Planning for Multi-Hop Question Answering}
\author{
    Yijie Zhu\equalcontrib\textsuperscript{\rm 1},
    Haojie Zhou\equalcontrib\textsuperscript{\rm 1},
    Wanting Hong\textsuperscript{\rm 1},
    Tailin Liu\textsuperscript{\rm 1},
    Ning Wang\thanks{Corresponding author.}\textsuperscript{\rm 1}
}
\affiliations{
    \textsuperscript{\rm 1}The School of Artificial Intelligence and Computer Science, Jiangnan University \\
    ningwang@jiangnan.edu.cn
}

\usepackage{bibentry}
\begin{document}

\maketitle

\begin{abstract}
Retrieval-augmented generation (RAG) has been extensively employed to mitigate hallucinations in large language models (LLMs). However, existing methods for multi-hop reasoning tasks often lack global planning, increasing the risk of falling into local reasoning impasses. Insufficient exploitation of retrieved content and the neglect of latent clues fail to ensure the accuracy of reasoning outcomes. To overcome these limitations, we propose \textbf{R}ecursive \textbf{E}valuation and \textbf{A}daptive \textbf{P}lanning (REAP), whose core idea is to explicitly maintain structured sub-tasks and facts related to the current task through the Sub-task Planner (SP) and Fact Extractor (FE) modules. SP maintains a global perspective, guiding the overall reasoning direction and evaluating the task state based on the outcomes of FE, enabling dynamic optimization of the task-solving trajectory. FE performs fine-grained analysis over retrieved content to extract reliable answers and clues. These two modules incrementally enrich a logically coherent representation of global knowledge, enhancing the reliability and the traceability of the reasoning process. Furthermore, we propose a unified task paradigm design that enables effective multi-task fine-tuning, significantly enhancing SP's performance on complex, data-scarce tasks. We conduct extensive experiments on multiple public multi-hop datasets, and the results demonstrate that our method significantly outperforms existing RAG methods in both in-domain and out-of-domain settings, validating its effectiveness in complex multi-hop reasoning tasks.

\end{abstract}

\begin{links}
    \link{Code}{https://github.com/Deus-Glen/REAP}
    % \link{Datasets}{https://aaai.org/example/datasets}
    % \link{Extended version}{https://aaai.org/example/extended-version}
\end{links}

\section{Introduction}
% hwt delete1
Large Language Models (LLMs) have demonstrated advanced capabilities across various natural language processing (NLP) tasks \cite{touvron2023llama, fan2024survey, gao2025synergizing}. However, their reliance on parameterized knowledge renders them prone to factually incorrect answers due to outdated information or hallucinations \cite{huang2025survey, cheng2024small}. To mitigate these limitations, Retrieval-augmented generation (RAG) has emerged as an effective approach for enhancing LLM performance in knowledge-intensive tasks by dynamically incorporating external non-parameterized knowledge sources \cite{lewis2020retrieval, gao2023retrieval}. While traditional RAG systems can efficiently handle simple single-hop question-answering (QA) tasks through a single retrieval-generation paradigm \cite{ye2024boosting, roy2024learning}, they struggle with multi-hop question-answering (MHQA) that require integrating information scattered across multiple documents \cite{he2024retrieving, trivedi2022interleaving, jiang2023active}.

% hwt delete4 + delete2 + delete3 
To overcome the challenges of MHQA, recent studies have adopted multi-round retrieval and iterative refinement strategies to progressively gather and filter relevant information. These approaches effectively integrate knowledge from multiple sources, significantly strengthening the reasoning capabilities of LLMs in MHQA tasks \cite{tang2024multihop, yang2024large, teng2025atom}. Nevertheless, such methods often suffer from inefficiencies in reasoning trajectory planning and limited depth of information exploitation. To identify optimal reasoning trajectories, some studies have introduced complex search algorithms such as Monte Carlo Tree Search (MCTS) \cite{dong2024progressive, li2024can}, but these typically come with significant computational overhead. Moreover, when solving sub-queries, models tend to extract direct answers, ignoring latent clues that are crucial to the final solution \cite{ye2025optimizing, wang2024sg}. Furthermore, to reinforce end-to-end reliability, recent efforts have introduced external components such as rerankers and decision-makers to improve retrieval accuracy \cite{glass2022re2g, jeong2024adaptive}, but such approaches often come at the expense of increased system complexity and diminished interpretability \cite{su2025fast}.

% hwt
To address the above limitations, we propose \textbf{R}ecursive \textbf{E}valuation and \textbf{A}daptive \textbf{P}lanning (REAP), a novel iterative framework to enhance the performance of RAG. Our core insight is that the incremental decomposition of complex queries into sub-queries may lead to reasoning impasses, while missing or inaccurate facts can cause deviations from the correct reasoning trajectory \cite{zhang2025credible}. To maintain coherent and reliable reasoning throughout the process, we explicitly maintain two critical knowledge sets: structured sub-tasks and facts, which serve as the foundation for guiding reasoning. Specifically, we introduce the Sub-Task Planner (SP) module that offers a global perspective, plans intermediate reasoning steps and dynamically guides the overall reasoning trajectory. Then, the Fact Extractor (FE) module is proposed to perform retrieval, integrate retrieved content with the structured facts for reasoning, thereby providing precise and complete facts and continuously expanding global context. These two modules are functionally decoupled yet tightly coordinated, continuously advancing the reasoning process. This modular and synergistic design allows the framework to rapidly identify optimal reasoning trajectories in complex scenarios while ensuring the reliability of facts, thereby enhancing the overall framework's performance.

Our contributions are summarized as follows:

\begin{enumerate}
    \item We propose REAP, a dual-module framework where the globally-aware SP module dynamically guides the reasoning trajectory, while the FE module continuously enriches the knowledge base with reliable facts, creating a mutually reinforcing cycle that enhances reasoning capabilities.
    \item We introduce a unified task paradigm that facilitates knowledge transfer from data-rich routine planning to data-scarce critical replanning via multi-task fine-tuning, significantly enhancing the framework's robustness and strategic intelligence.
    \item We separate simple and complex reasoning into distinct sub-modules for SP, enabling the simple sub-module to be replaced with a more lightweight model to improve efficiency while preserving performance.
    \item We conduct extensive experiments on multiple MHQA datasets, demonstrating the effectiveness and robustness of REAP.
\end{enumerate}

\section{Related Work}
\subsection{RAG}
% hwt delete5
Traditional RAG frameworks have been introduced to mitigate issues such as knowledge obsolescence and hallucinations in LLMs by incorporating external knowledge sources. Early representative methods like Standard RAG \cite{lewis2020retrieval} follow a static retrieve-then-generate paradigm, typically leveraging either sparse retrieval \cite{robertson2009probabilistic} or dense retrieval \cite{karpukhin2020dense, wang2022text} to refine the factual correctness of generated responses. Although these methods have promised accuracy of QA, their effectiveness in complex reasoning tasks remains limited, particularly in multi-hop reasoning tasks due to the lack of dynamic interaction and reasoning mechanisms. To address the above challenges, researchers have explored refinements to single-round RAG from multiple perspectives, aiming to better support complex reasoning tasks. Prior to retrieval, query enhancement and rewriting are used to optimize information alignment \cite{gao2023precise, ma2023query, chan2024rq}. During the retrieval process, dynamic strategies guided by confidence intervals have been introduced to reduce redundant computations and improve retrieval efficiency \cite{jiang2023active, su2024dragin, DBLP:conf/aaai/Deng0ZWF25}. Following retrieval, the retrieved content is further refined through end-to-end training \cite{shi2023replug} or post-processing modules \cite{glass2022re2g, jiang2023llmlingua,kim2024sure} to enhance generation quality. Despite these advancements, the inherent inadequacy of traditional RAG with single-round retrieval in handling complex, multi-hop reasoning tasks has prompted a focus on iterative retrieval-generation architectures.
\subsection{Iterative RAG}
% hwt delete3 + delete6
Iterative RAG frameworks operate by interleaving retrieval and generation across multiple rounds to incrementally synthesize evidence and construct reasoning chains in a context-sensitive and adaptive manner \cite{ram2023context, yang2024rag}. To achieve this, A fundamental strategy is problem decomposition, which partitions complex questions into simpler sub-queries, retrieves supporting evidence for each sub-query, and subsequently integrates the sub-answers to facilitate multi-hop reasoning \cite{xu2024search, shi2024generate}. To further advance the reasoning planning process, a number of studies introduce search-based algorithms, such as MCTS, to simulate and evaluate alternative reasoning branches and identify optimal reasoning strategies \cite{dong2024progressive, li2024can}. Another line of research integrates chain-of-thought prompting with multi-round retrieval to guide the model in progressively constructing the final answer \cite{yao2023react, press2022measuring, wang2025chain}. Representative methods such as IRCoT \cite{trivedi2022interleaving} and Iter-RetGen \cite{shao2023enhancing} dynamically generate sub-queries, retrieve relevant documents, and iteratively refine the reasoning trajectory throughout the generation process. Additionally, to strengthen strategic control capabilities during multi-round retrieval, some methods incorporate self-reflection mechanisms, enabling the model to autonomously determine whether to continue retrieving or proceed with answer generation \cite{asai2023self}, while others rely on reinforcement learning frameworks to learn retrieval and generation policies based on interactive feedback \cite{jin2025search, song2025r1}. However, these methods often fall short in global planning and facts extraction, leading to brittle and incoherent reasoning trajectories. In contrast to the aforementioned methods, our proposed REAP substantially enhances reasoning ability and robustness in multi-hop reasoning tasks through a recursive feedback loop between SP and FE.

\section{Preliminary}
\label{sec:preliminary}

\begin{figure*}[htbp]
  \centering 
  \includegraphics[width=\textwidth]{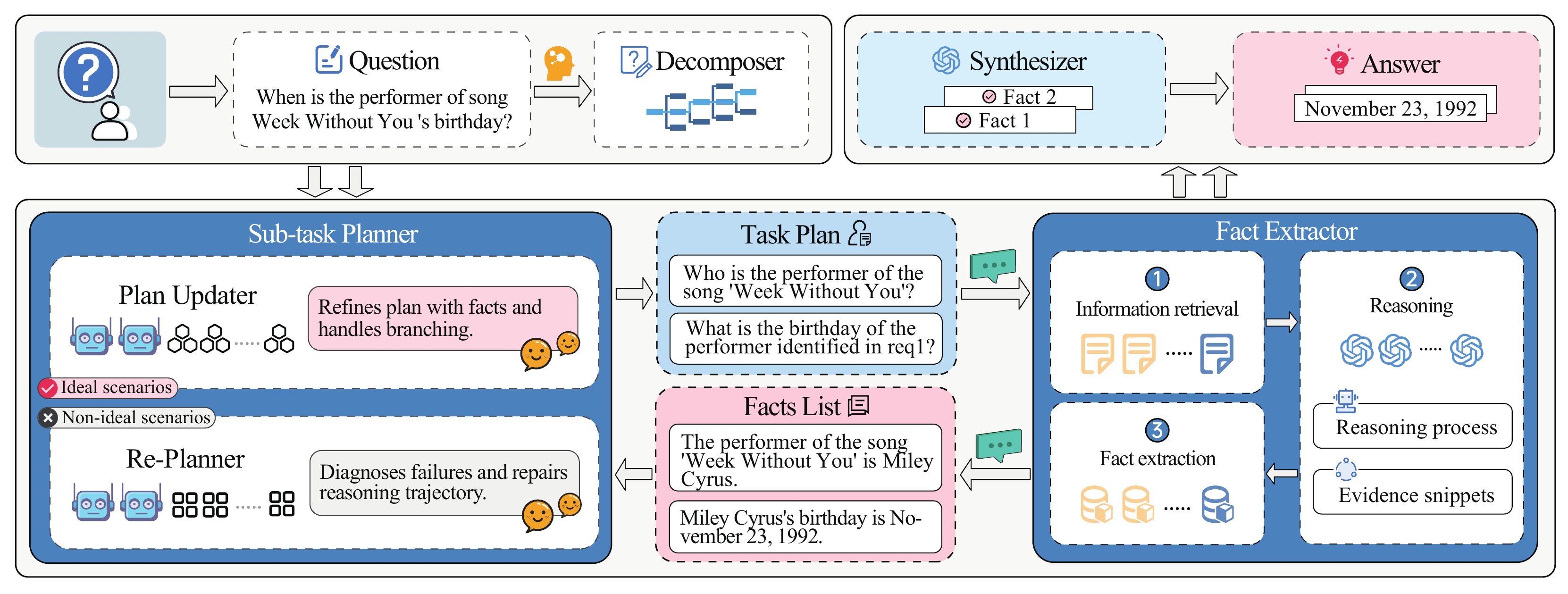} 
  \caption{Overall framework of REAP. After the Decomposer breaks down the question, the Sub-task Planner and Fact Extractor iterate to guide the reasoning trajectory and gather the facts, and the Synthesizer produces the final answer.}
  \label{figure1}
\end{figure*}

For multi-hop scenarios, we formally define the iterative RAG task as follows: Given a query $Q$ and an external corpus $\mathcal{C}$, the objective is to generate a factual answer $A$. This task is typically operationalized as an iterative reasoning process, where at each step $t$, a sub-query $q_t$ is formulated to gather a specific piece of information. A retriever function $\text{Retriever}(\cdot)$, then fetches a set of relevant documents $D_t \subset \mathcal{C}$ based on the sub-query:

\begin{equation}
    D_t = \text{Retriever}(q_t; \mathcal{C})
    \label{eq:retriever}
\end{equation}

An LLM, $M_\theta$ parameterized by $\theta$, processes these documents, often conditioned on the history of previous interactions $H_{t-1} = \{Q, (q_1, A_1), \dots, (q_{t-1}, A_{t-1})\}$, to generate a sub-answer $A_t$:
\begin{equation}
    A_t = M_\theta(\text{GenerateSubAnswer} \mid Q, D_t, H_{t-1})
    \label{eq:generate}
\end{equation}

The history is then updated with the new pair $(q_t, A_t)$, and the process repeats until a termination condition is met. Finally, a synthesis function generates the final answer $A$ based on the complete interaction history $H_{\text{final}}$:
\begin{equation}
    A = M_\theta(\text{Synthesize} \mid Q, H_{\text{final}})
    \label{eq:synthesize}
\end{equation}

This iterative paradigm allows the model to progressively build a chain of reasoning by decomposing the original complex query into a sequence of sub-queries.

\section{Method}

\subsection{Framework Overview}

Our proposed framework, REAP, reframes the multi-hop reasoning process from a linear pipeline into a dynamic, state-driven loop. The architecture, illustrated in Figure \ref{figure1}, is designed around the core principle of explicitly decoupling planning from fact-gathering to enhance robustness and interpretability. It primarily consists of two synergistic modules operating in a recursive loop: the Sub-task Planner (SP) and the Fact Extractor (FE).

The overall workflow proceeds as follows:
Given a complex query $Q$, an initial Decomposer first generates a structured task plan, $\mathcal{P}_0 = \{p_1, p_2, \dots, p_N\}$. Each sub-task $p_i \in \mathcal{P}_0$ is a tuple $(id_i, q_i, deps_i)$, representing its unique identifier, query, and dependencies. This plan, along with an initially empty facts list $\mathcal{F}_0 = \emptyset$, defines the initial state of the framework.

The core of REAP is an iterative process that continues until a termination condition is met. At each step $t$:
\begin{enumerate}
    \item The SP, acting as the strategic core, analyzes the current state $(\mathcal{P}_{t-1}, \mathcal{F}_{t-1})$ to determine a set of executable next actions, $\text{Actions}_t$.
    \item For each sub-task $p_i \in \text{Actions}_t$, the FE performs retrieval and grounded reasoning to produce a new fact object, $f_i$.
    \item The facts list is updated: $\mathcal{F}_t = \mathcal{F}_{t-1} \cup \{f_1, f_2, \dots\}$. The SP then receives this feedback to generate the subsequent plan, $\mathcal{P}_t$.
\end{enumerate}

This recursive loop, where the SP guides the FE and the FE's findings inform the SP's next planning cycle, continues until the plan is fully resolved. Finally, a Synthesizer generates the conclusive answer $A$ by reasoning over the final, comprehensive facts list $\mathcal{F}_{final}$ and the original query $Q$:
\begin{equation}
    A = M_\theta(\text{Synthesize} \mid Q, \mathcal{F}_{final})
\end{equation}
This modular and iterative design allows REAP to navigate complex reasoning trajectories, recover from errors, and construct a fully traceable line of evidence.

\subsection{Sub-task Planner (SP)}

The SP serves as the strategic core of the REAP framework, designed to overcome the limitations of myopic, step-by-step reasoning that often leads to local impasses or deadlocks. By receiving and dynamically maintaining the complete initial task plan $\mathcal{P}_0$, the SP retains a global perspective throughout the process. A key innovation of the SP is its state-aware, modular design. Based on the fulfillment level $l_t$ of the new fact $f_t$ returned by the FE, the SP dispatches the task to one of two specialized sub-modules:

\begin{enumerate}
    \item \textbf{Plan Updater}: This sub-module handles ideal scenarios where reasoning progresses as expected (e.g., $l_t$ is \texttt{DirectAnswer}). It performs deterministic and rule-based updates to maintain the task plan. Its primary functions are:
    \begin{itemize}
        \item \textit{Fact Substitution}: It systematically rewrites pending sub-tasks by substituting abstract placeholders with concrete entities derived from newly acquired facts. This ensures that subsequent queries are self-contained and fully grounded.
        \item \textit{Plan Forking}: In cases where a sub-task yields a set of multiple valid sub-answers, this function programmatically duplicates the subsequent dependent sub-tasks into parallel branches, ensuring all valid reasoning trajectories are exhaustively explored.
    \end{itemize}

    \item \textbf{Re-Planner}: This sub-module is invoked to handle non-ideal scenarios (e.g., $l_t$ is \texttt{PartialClue} or \texttt{Failed}), acting as the critical reasoning and recovery mechanism. Its hierarchical responsibilities are:
    \begin{itemize}
        \item \textit{Pragmatic Sufficiency Assessment}: Its first and most crucial task is to evaluate whether a partially fulfilled sub-task, despite its incompleteness, is functionally sufficient to satisfy the informational requirements of subsequent reasoning steps. This assessment is made by analyzing the dependencies of the downstream plan relative to the ultimate query $Q$. If the partial information is deemed sufficient, the sub-task is considered resolved, thus preventing inefficient, perfectionist search loops.
        \item \textit{Scoped Plan Repair}: Only if the acquired information is assessed as insufficient does the Re-Planner proceed with plan repair. It first diagnoses the failure's scope---differentiating between a localized issue (e.g., a poorly formulated query) and a systemic flaw (e.g., an irrelevant reasoning trajectory). For localized issues, it performs a minor adjustment by refining the sub-task's query. For systemic flaws, it executes a major overhaul by pruning the invalid branch and injecting a new, more logical sequence of sub-tasks to create an alternative solution path.
    \end{itemize}
\end{enumerate}

At an iterative step $t$, given the task plan $\mathcal{P}_{t-1}$ and facts list $\mathcal{F}_{t-1}$ from the previous state, along with the new fact $f_t$, the decision process of the SP can be formalized as:
\begin{equation}
    (\mathcal{P}_t, \text{Actions}_t) = \text{SP}( \mathcal{P}_{t-1}, \mathcal{F}_{t-1} \cup \{f_t\}, Q )
\end{equation}
where $\text{Actions}_t$ is the list of concrete sub-tasks to be executed next. The internal dispatch logic of the SP is determined by the fulfillment level $l_t$ of the new fact $f_t$:
\begin{equation}
\text{SP} \leftarrow
\begin{cases}
\text{Plan Updater} & \text{if } l_t = \text{\texttt{DirectAnswer}} \\
\text{Re-Planner} & \text{otherwise}
\end{cases}
\end{equation}
This dual-module design allows REAP to handle routine progress with high efficiency while possessing robust, intelligent capabilities to recover from complex failures and adapt its strategy in real-time.

\subsection{Fact Extractor (FE)}

The facts list $\mathcal{F}$ serves as the evolving knowledge foundation for the entire reasoning process; thus, its reliability and comprehensiveness are paramount. The Fact Extractor module is designed to extract high-fidelity, structured facts from retrieved documents. To mitigate hallucinations and enhance traceability, we require the LLM to not only provide a conclusive statement but also articulate its reasoning process and cite direct textual evidence.

A key function of the FE is its ability to discern not just direct answers but also latent clues that may be crucial for subsequent reasoning steps. For a given sub-query $q_t$ from a sub-task $p_t \in \text{Actions}_t$, the FE process is defined as follows:

First, a retriever function fetches a set of relevant documents $D_t$ from an external corpus $\mathcal{C}$.
Next, an LLM $M_\theta$ processes these documents to generate a new structured fact object $f_t$. Crucially, the model is conditioned not only on the current query and retrieved documents but also on the historical facts $\mathcal{F}_{t-1}$. This contextual conditioning allows the model to perform more sophisticated reasoning, such as co-reference resolution and identifying relationships between new information and previously established facts. The generation process is formulated as:
\begin{equation}
    f_t = M_\theta(\text{ExtractFact} | q_t, D_t, \mathcal{F}_{t-1})
\end{equation}
The fact object $f_t$ is a structured tuple designed to capture the richness of the extracted information:
\begin{equation}
    f_t = (s_t, e_t, r_t, l_t)
\end{equation}
where:
\begin{itemize}
    \item $s_t$ is the core statement, a concise, self-contained factual assertion.
    \item $e_t \subseteq D_t$ is the set of direct textual evidence snippets that ground the statement $s_t$.
    \item $r_t$ is the model's reasoning process, a chain-of-thought explanation of how $s_t$ is derived from $e_t$.
    \item $l_t$ is the fulfillment level, a categorical label that classifies the quality of the extracted fact. This level is crucial for the SP's subsequent decision-making. 
\end{itemize}
The statement $s_t$ and fulfillment level $l_t$ are then used to directly guide the SP module's next planning cycle.

\subsection{Multi-task Fine-tuning}

To enhance the performance of the planning-related modules within the REAP framework, especially for the Re-Planner, which suffers from data scarcity due to its lower invocation frequency, we devise a multi-task fine-tuning strategy.

The core insight is that despite their varying difficulty, the Decomposer, Plan Updater, and Re-Planner modules share a significant functional commonality: they all require the model to generate or modify a structured task plan based on existing information. We leverage this commonality by consolidating their respective datasets, $\mathcal{D}_{\text{decomp}}, \mathcal{D}_{\text{update}}, \mathcal{D}_{\text{replan}}$, for joint training of a single planning model $M_\phi$.

The training objective is to minimize a weighted joint loss function $\mathcal{L}_{\text{multi}}$:
{
\fontsize{8}{12}\selectfont
\begin{equation}
    \min_{\phi} \mathcal{L}_{\text{multi}}(\phi) = \sum_{task \in \mathcal{T}} \lambda_{task} \mathbb{E}_{(x, y) \sim \mathcal{D}_{task}} [ \mathcal{L}_{\text{task}}(M_\phi(x), y) ]
\end{equation}
}
where $\mathcal{T} = \{\text{decomp, update, replan}\}$, $\lambda_{task}$ is the task-specific weight, set to 1, and $\mathcal{L}_{\text{task}}$ is the corresponding loss function. This paradigm enables effective knowledge transfer from data-rich tasks (Decomposer, Plan Updater) to the data-scarce Re-Planner task. This significantly enhances the robustness and accuracy of the Re-Planner in complex, anomalous scenarios, thereby elevating the overall intelligence of the REAP framework.
\begin{table*}[htbp]
\centering
\begin{tabular}{l ccc ccc ccc ccc} 
\toprule
\multirow{2}{*}{\textbf{Models}}
& \multicolumn{3}{c}{\textbf{HotpotQA}} & \multicolumn{3}{c}{\textbf{2Wiki}} & \multicolumn{3}{c}{\textbf{MuSiQue}\textsuperscript{‡}} & \multicolumn{3}{c}{\textbf{Bamboogle}\textsuperscript{‡}} \\ 
\cmidrule(lr){2-4} \cmidrule(lr){5-7} \cmidrule(lr){8-10} \cmidrule(lr){11-13} 
& CEM & F1 & ACC\textsuperscript{†} & CEM & F1 & ACC\textsuperscript{†} & CEM & F1 & ACC\textsuperscript{†} & CEM & F1 & ACC\textsuperscript{†} \\ 
\midrule
Naive & 24.6  & 31.8  & 34.6  & 31.6  & 35.1  & 32.8  & 6.2  & 10.6  & 10.0  & 13.6  & 17.7  & 21.6  \\
Standard RAG & 41.6  & 48.6  & 51.2  & 36.2  & 38.5  & 38.2  & 9.8  & 13.5  & 12.6  & 20.8  & 30.9  & 32.0  \\
FT-Standard RAG & 53.2  & 61.2  & 65.4  & 53.6  & 57.1  & 56.2  & 18.2  & 26.4  & 22.6  & 42.4  & 54.4  & 51.2 \\
SuRe & 29.4  & 33.8  & 40.4  & 24.4  & 24.6  & 28.2  & 5.8  & 11.6  & 11.4  & 10.4  & 19.1  & 20.0  \\
IRCoT & 42.2  & 51.4  & 54.8  & 41.2  & 36.5  & 41.8  & 18.2  & 18.6  & 20.4  & 35.2  & 30.1  & 44.8  \\
Iter-Retgen & 45.0  & 43.0  & 54.6  & 42.0  & 32.2  & 41.0  & 14.2  & 18.6  & 20.6  & 19.2  & 27.2  & 28.0  \\
SearChain & 39.8  & 39.6  & 50.4  & 50.6  & 46.7  & 50.0  & 20.6  & 23.6  & 26.0  & 47.2  & 49.4  & 53.6 \\
Search-R1 & 44.8  & 52.7  & 57.0  & 47.4  & 49.7  & 50.8  & 20.0  & 25.7  & 25.8  & 40.8  & 50.8  & 49.6  \\
R1-Searcher & \underline{56.8}  & \underline{63.4}  & \underline{68.4}  & \underline{68.4}  & \underline{69.4}  & \underline{70.0}  & \underline{29.2}  & \underline{33.8}  & \underline{34.4}  & \underline{48.8}  & \underline{58.0}  & \underline{56.0}  \\ 
\midrule
REAP & \textbf{59.2}  & \textbf{68.0}  & \textbf{72.4}  & \textbf{79.2}  & \textbf{79.6}  & \textbf{81.6}  & \textbf{33.6}  & \textbf{38.3}  & \textbf{40.8}  & \textbf{49.6}  & \textbf{65.2}  & \textbf{65.6}  \\ 
\bottomrule
\end{tabular}
\caption{Performance comparisons between REAP and the baselines on four MHQA benchmarks. Bold numbers indicate the best result, while underlined number indicates the second-best result. ‡ represents out-of-domain datasets.}
\label{table1}
\end{table*}

\section{Experiments}
\subsection{Datasets and Metrics}
We conduct evaluations of our method on four multi-hop datasets: HotpotQA \cite{yang2018hotpotqa}, 2WikiMultihopQA \cite{ho2020constructing}, MuSiQue \cite{trivedi2022musique}, and Bamboogle \cite{press2022measuring}. HotpotQA and 2WikiMultihopQA are in-domain benchmarks, as parts of their training sets are used for model training, whereas MuSiQue and Bamboogle serve as out-of-domain benchmarks to assess the generalization capabilities of our method. For the first three datasets, we randomly sample 500 examples from each validation split as test sets. For Bamboogle, since it only contains 125 examples, we use the entire test set for evaluation. Regarding evaluation metrics, we adopt Cover Exact Match (CEM), F1 score, and ACC\textsuperscript{†} (with an LLM serving as the judge). Detailed datasets and metrics descriptions are provided in the Appendix.
\subsection{Baselines}
% hwt
We compare REAP against the following baselines. The Naive method directly generates answer without any retrieval. The Standard RAG \cite{lewis2020retrieval} performs one-step retrieval and concatenates the documents with the question for answer generation. Additionally, we fine-tune the Standard RAG with the same training data adopted in our method, and name it FT-Standard RAG. SuRe \cite{kim2024sure} ranks answer candidates by generating conditioned summaries of the retrieved content. IRCoT \cite{trivedi2022interleaving}, Iter-RetGen \cite{shao2023enhancing} and SearChain \cite{xu2024search} combine reasoning chains with multi-round retrieval to form the reasoning trajectory. Search-R1 \cite{jin2025search} and R1-Searcher \cite{song2025r1} employ template-guided prompting, reinforcement learning, and summarization mechanisms to achieve more refined integration of evidence and answer generation.
% R1-Searcher obtains superior performance compared to other baselines.
\subsection{Implementation Details}
In our experiments, REAP and other non-fine-tuned baseline methods use Llama-3.1-8B-Instruct \cite{grattafiori2024llama} as the generator and are evaluated with FlashRAG \cite{jin2025flashrag}. For methods involving fine-tuning, we use the model checkpoints provided by the authors. We adopt the corpus provided by CoRAG \cite{wang2025chain}, which is based on the English Wikipedia provided by KILT, containing approximately 36 million passages. We use e5-large-v2 \cite{wang2022text} as the main retriever, with the top-5 results returned for each query. For all multi-round methods, we set the maximum number of iterations to 5. We randomly select 7,000 samples from HotpotQA and WikiMultihopQA and run REAP using GPT-4 \cite{achiam2023gpt} to collect training data. After filtering, we finally select 5,556 samples as the training set, of which 2,988 are from HotpotQA and 2,568 are from 2WikiMultihopQA. Detailed training setting is provided in the Appendix.
\subsection{Main Results}
Table \ref{table1} presents the experimental results of REAP and other baseline methods on four representative MHQA datasets. Specifically, we observe the following:
\begin{enumerate}
\item Iterative interaction leads to substantial performance gains. Compared with the Standard RAG method, REAP improves the F1 score on HotpotQA from 48.6\% to 68.0\%, demonstrating the superiority of the iterative approach for solving MHQA. More importantly, REAP maintains a significant advantage over the FT-Standard RAG, achieving a 6.8\% improvement in F1 score, indicating that the performance gain is not merely attributable to the training data but primarily stems from our proposed iterative reasoning and evidence extraction framework.
\item REAP significantly outperforms existing multi-round methods. It impressively surpasses the top-performing R1-Searcher method in FlashRAG on all datasets, achieving F1 score improvements of 4.6\% on HotpotQA and 10.2\% on 2WikiMultihopQA. This demonstrates that our method effectively enhances the model's ability to reason and extract factual evidence.

\begin{table}[htbp] 
\centering
\begin{tabular}{l c c c c} 
\toprule
\multirow{2}{*}{\textbf{Models}}
& \multicolumn{1}{c}{\textbf{HQA}} & \multicolumn{1}{c}{\textbf{WQA}} & \multicolumn{1}{c}{\textbf{MQA}\textsuperscript{‡}} & \multicolumn{1}{c}{\textbf{BQA}\textsuperscript{‡}} \\
\cmidrule(lr){2-5}
& F1  & F1  & F1 & F1  \\ 
\midrule
w/o replan   & 64.9    & 78.6   & 34.2  & 61.6 \\ 
w/o verify   & 65.1    & 78.0   & 34.8  & 60.8\\ 
w/o clue     & 64.6    & 76.5   & 35.2  & 62.7\\ 
REAP         & 68.0    & 79.6   & 38.3  & 65.2\\
\bottomrule
\end{tabular}
\caption{The ablation study on HotpotQA (HQA), 2WikimultihopQA (WQA), MuSiQue (MQA) and Bamboogle (BQA).}
\label{table2}
\end{table}

\item The model exhibits strong generalization capability. REAP is trained on only 5,556 samples from HotpotQA and 2WikiMultihopQA. It not only excels on these in-domain datasets but also achieves the best scores on out-of-domain datasets, showcasing its powerful generalization. This suggests that our training process enables the model to learn the fundamental skills of reasoning and facts extraction, rather than simply memorizing a generation pattern, thus leading to robust performance on unseen data.
\end{enumerate}

\section{Analysis}
\subsection{Ablation Studies}
We conduct ablation studies on the same four multi-hop datasets to analyze the effectiveness of key module designs within our proposed REAP framework. Meanwhile, we also investigate the contribution of our multi-task fine-tuning strategy to the framework.
\subsubsection{Module Ablation}
To evaluate the effectiveness of our individual mechanisms, we disable key functionalities from the modules. For the SP module, we ablate the Re-Planner sub-module (w/o replan), which is designed to handle challenging scenarios. For the FE module, we ablate the logical reasoning and verification mechanism (w/o verify) and the clue feedback mechanism (w/o clue), respectively. The results are presented in Table \ref{table2}.

We observe that the full REAP framework consistently outperforms all ablated variants, confirming the positive contribution of each mechanism. The most significant performance degradation is observed in w/o replan, with F1 score dropping by 3.1\% on HQA, 4.1\% on MQA, and 3.6\% on BQA. This substantial decline, particularly on the more complex MQA and BQA datasets, underscores the criticality of the Re-Planner's ability to dynamically correct and reroute the reasoning trajectory when encountering failures or suboptimal steps.

Removing the verification mechanism (w/o verify) consistently lowers performance, with a particularly sharp drop of 4.4\% F1 score on BQA. This highlights the necessity of ensuring factual accuracy at each iterative step. Without robust verification, incorrect or hallucinatory information can be propagated, leading to a cascade of errors that derail the entire reasoning trajectory.

\begin{figure}[htbp]
  \centering 
  \includegraphics[width=\columnwidth]{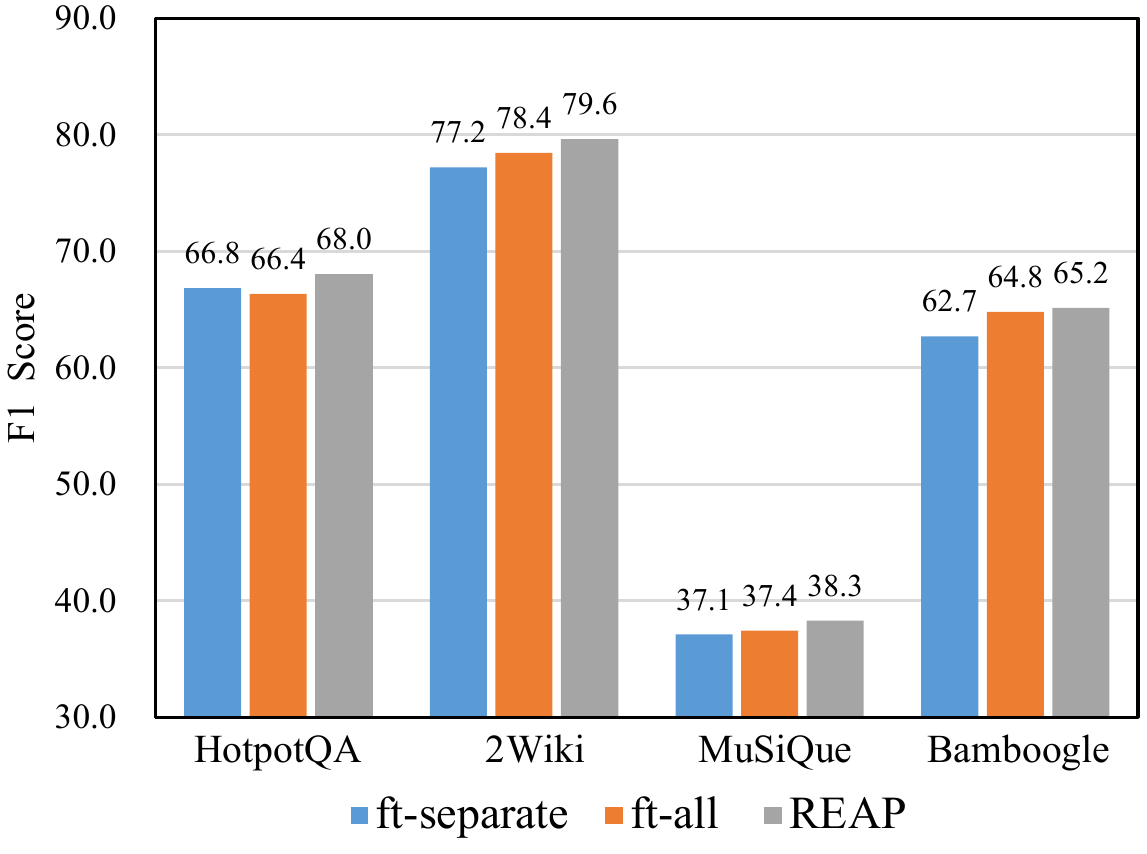}
  \caption{Ablation study on the effectiveness of Multi-task Fine-tuning.}
  \label{figure2}
\end{figure}

Disabling the clue feedback mechanism (w/o clue) also results in a notable performance drop across all benchmarks, for instance, a decrease of 3.4\% F1 score on HQA and 3.1\% on WQA. This demonstrates that the model's capacity to identify and leverage partial or latent clues, beyond just direct answers, is vital for constructing a comprehensive factual basis required for the final answer synthesis.

\subsubsection{Multi-task Fine-tuning Ablation}
To validate the effectiveness of our proposed multi-task fine-tuning strategy, we conduct an ablation study. We compare REAP with two settings: ft-separate, where we fine-tune a dedicated model for each module (Decomposer, Plan Updater, and Re-Planner) using only its corresponding training data; and ft-all, where a single model is jointly fine-tuned on the combined data from all modules.

\begin{table*}[htbp] 
\centering
\begin{tabular}{l ccc ccc ccc ccc} 
\toprule
\multirow{2}{*}{\textbf{Models}}
& \multicolumn{3}{c}{\textbf{HotpotQA}} & \multicolumn{3}{c}{\textbf{2Wiki}} & \multicolumn{3}{c}{\textbf{MuSiQue}} & \multicolumn{3}{c}{\textbf{Bamboogle}} \\ 
\cmidrule(lr){2-4} \cmidrule(lr){5-7} \cmidrule(lr){8-10} \cmidrule(lr){11-13} 
& CEM & F1 & ACC\textsuperscript{†} & CEM & F1 & ACC\textsuperscript{†} & CEM & F1 & ACC\textsuperscript{†} & CEM & F1 & ACC\textsuperscript{†} \\ 
\midrule
Naive\textsubscript{70B} & 32.4  & 39.7  & 40.4  & 36.8  & 39.3  & 38.8  & 11.6  & 15.1  & 17.4  & 37.6  & 37.0  & 44.0  \\ 
StandardRAG\textsubscript{70B} & 51.0  & \underline{56.3}  & 61.2  & 46.8  & \underline{47.2}  & 46.6  & 16.8  & 21.1  & \underline{24.8}  & 35.2  & 46.1  & 46.4  \\ 
SuRe\textsubscript{70B} & 35.8  & 43.1  & 46.0  & 27.6  & 34.6  & 33.2  & 12.2  & 20.4  & 17.4  & 10.4  & 19.1  & 20.0  \\ 
IRCoT\textsubscript{70B} & \underline{55.8}  & 49.5  & 60.4  & \underline{57.0}  & 33.1  & 35.8  & \underline{21.0}  & 19.4  & 24.3  & 32.0  & 38.8  & 40.8  \\
Iter-RetGen\textsubscript{70B} & 53.8  & 47.6  & \underline{65.0}  & 51.4  & 41.5  & \underline{53.6} & 17.2  & \underline{21.6}  & 24.6  & \underline{41.6}  & \underline{47.5}  & \underline{52.8}  \\
\midrule
REAP\textsubscript{70B} & \textbf{63.2}  & \textbf{65.5}  & \textbf{73.6}  & \textbf{73.8}  & \textbf{70.0}  & \textbf{72.2}  & \textbf{37.6}  & \textbf{37.2}  & \textbf{42.6}  & \textbf{54.4}  & \textbf{61.6}  & \textbf{63.2} \\
\bottomrule
\end{tabular}
\caption{Performance comparison of REAP and non-fine-tuned baselines on four MHQA benchmarks under the Llama-3.1-70B-Instruct setting. Bold numbers indicate the best result, while underlined number indicates the second-best result.}
\label{table3}
\end{table*}

The results, presented in Figure \ref{figure2}, reveal the clear advantages of our method. The ft-separate setting consistently yields the lowest performance, lagging behind REAP by a significant margin of 2.5\% F1 score on Bamboogle and 2.4\% on 2Wiki. This demonstrates that training on isolated tasks prevents the model from learning the underlying correlations and generalizable reasoning patterns shared across the planning modules, thereby limiting its capabilities.

Crucially, our REAP framework, which employs the carefully designed multi-task fine-tuning strategy, consistently achieves the highest scores across all benchmarks. It surpasses the ft-all setting by 1.2\% F1 score on 2Wiki and 0.9\% on MuSiQue. This final improvement highlights that merely combining training data is insufficient. Our multi-task fine-tuning strategy more effectively facilitates the transfer of robust planning capabilities to the data-scarce but critical Re-Planner module, confirming the effectiveness and rationale of our methodology.

\begin{figure}[htbp]
  \centering 
  \includegraphics[width=\columnwidth]{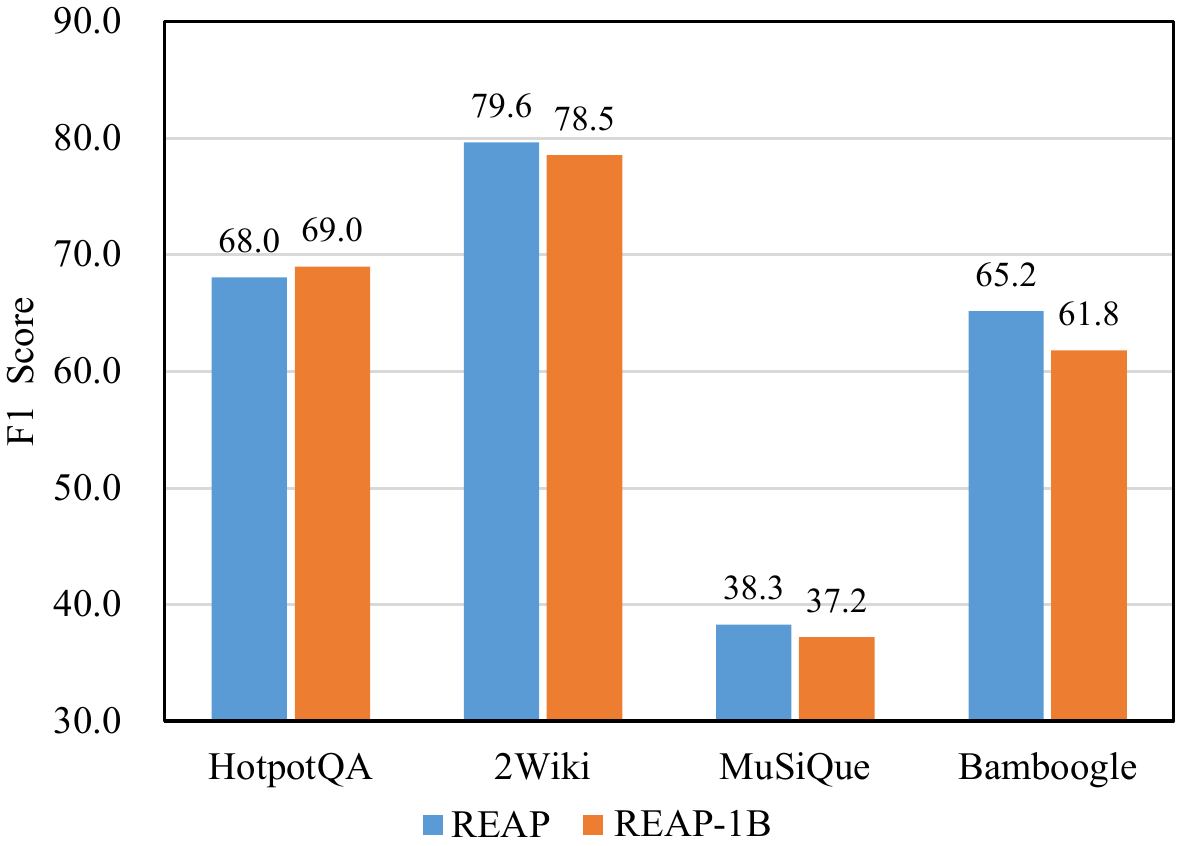}
  \caption{Module substitution experiment validating the efficiency gains of asymmetric configuration.}
  \label{figure3}
\end{figure}

\subsection{Further Analysis}
\subsubsection{Efficiency Analysis}
In practical applications, in addition to pursuing the highest performance, inference efficiency (i.e., latency and computational cost) is also a key consideration. Our REAP framework dynamically assigns simple and complex scenarios to different modules. The Plan Updater sub-module for simple cases is relatively simple, and according to our statistics, about 85\% of steps are classified as simple scenarios. To improve overall efficiency, we replace the base model in the Plan Updater sub-module with Llama-3.2-1B-Instruct, fine-tuned on the same data. As shown in Figure \ref{figure3}, even with an 1B-parameter model, the task can still be accomplished with relatively high accuracy. At the same time, the framework benefits from the faster inference speed of the smaller model, resulting in improved overall efficiency. This discovery fully demonstrates the unique advantages of the REAP framework's asymmetric modularity design: it allows us to flexibly configure different scales of computing resources for different modules, achieving a delicate balance between performance and efficiency.

\subsubsection{Performance on Larger Models}
To investigate the scalability of the REAP framework and its interaction with the capabilities of the base model, we conduct further experiments, replacing the base model from Llama-3.1-8B-Instruct with the more powerful Llama-3.1-70B-Instruct, and comparing its performance with five other non-fine-tuning methods. All methods are not fine-tuned to ensure fairness. As shown in Table \ref{table3}. We observe two key phenomena. First, when the base model is replaced with Llama-3.1-70B-Instruct, most methods achieve improved performance. For example, the F1 score of Standard RAG on HotpotQA increases from 39.7\% to 56.3\% compared to Naive, confirming the general benefit of employing more powerful LLMs for complex MHQA tasks. More importantly, our REAP framework retains its leading performance when built upon a stronger base model. This result provides compelling evidence for the superiority of the REAP framework, suggesting that its efficient iterative mechanism can serve as a capability amplifier: when the base model possesses stronger reasoning and generation ability, REAP can effectively guide and leverage these strengths without fine-tuning, thereby achieving further performance gains.

\subsection{Case Study}
To clearly demonstrate the workflow of the REAP framework, we present a case study that showcases the precise task decomposition by the Decomposer, the adaptive plan updates and action selection by the SP module, and the meticulous reasoning and verification performed by the FE module, all synergistically interacting to arrive at the final answer. A detailed example is provided in the Appendix.

\section{Conclusion}
To overcome the limitations of existing RAG methods in MHQA, where they often fall into local impasses or suffer from factual inaccuracies, we propose REAP, a novel framework that enhances reasoning reliability through iterative interaction. At the core of REAP are two synergistic modules: the Sub-task Planner (SP) and the Fact Extractor (FE). The SP maintains a global planning perspective to dynamically evaluate and optimize the reasoning trajectory, while the FE performs fine-grained analysis on retrieved content to extract high-fidelity facts. This decoupled yet tightly coordinated design ensures the coherence and accuracy of the reasoning process. Furthermore, by leveraging a multi-task fine-tuning paradigm and replacing sub-module with a more lightweight model, we improve performance while maintaining inference efficiency. Extensive experiments on multiple MHQA datasets demonstrate that REAP significantly outperforms existing state-of-the-art methods in both in-domain and out-of-domain settings, thereby validating its effectiveness and robustness in addressing complex multi-hop reasoning tasks.

\nocite{zheng2024llamafactory}
\nocite{zhu2025reap}

\bibliography{aaai2026}

\end{document}